\documentclass[runningheads]{llncs}
\usepackage[T1]{fontenc}
\usepackage{graphicx}
\usepackage{tcolorbox}
\usepackage{amsmath} 
\usepackage{amssymb} 
\usepackage{booktabs}

\usepackage{tikz}
\usepackage{array} 

\usetikzlibrary{positioning, arrows.meta, shapes.geometric, fit, calc, backgrounds, decorations.pathreplacing, patterns}

\definecolor{myteal}{RGB}{0, 128, 128}

\usepackage{pgfplots}
\pgfplotsset{compat=1.17} 

\usepackage{enumitem}
\usepackage{pifont}

\usepackage{microtype}

\begin{document}
\title{What Matters for Grocery Product Retrieval with Open Source Vision Language Models}
\titlerunning{Grocery Product Retrieval with Open Source VLMs}
%
\author{Emmanuel G. Maminta\inst{1} \and
Rowel O. Atienza\inst{1,2}}
\authorrunning{E. G. Maminta \and R. O. Atienza}
%
\institute{AI Graduate Program, University of the Philippines, Diliman, Quezon City \and
EEEI, University of the Philippines, Diliman, Quezon City \\
\email{emmanuel.maminta@eee.upd.edu.ph}, \email{rowel@eee.upd.edu.ph}}
\maketitle              
%


\begin{abstract}
Multimodal product retrieval (MPR) underpins checkout-free retail and automated inventory systems, yet it demands fine-grained SKU discrimination that standard vision-language benchmarks fail to capture. We present the first systematic zero-shot evaluation of 190 open-source VLMs on the MPR task of the GroceryVision Challenge, isolating pre-training data, architecture, and input resolution. Our analysis yields three actionable findings. \textbf{(1) Data quality trumps scale.} Switching from raw web-scrapes to filtered datasets delivers up to 16.6\% accuracy gains, exceeding the benefit of doubling model parameters. \textbf{(2) Efficient models can win.} MobileCLIP-B (150M parameters) outperforms 351M counterparts trained on noisy data. We introduce \textit{semantic power density} ($\phi$), an efficiency metric that penalizes sub-threshold accuracy. \textbf{(3) A precision gap persists.} State-of-the-art models achieve 94.5\% Recall@5 but suffer a 17.5\% drop at Recall@1, revealing that contrastive embeddings cluster categories effectively but fail to rank visually similar SKUs. Code and evaluation scripts are available at \url{https://github.com/upeee/openmpr}.

\keywords{Multimodal product retrieval \and VLMs \and Retail intelligence.}
\end{abstract}

\section{Introduction}

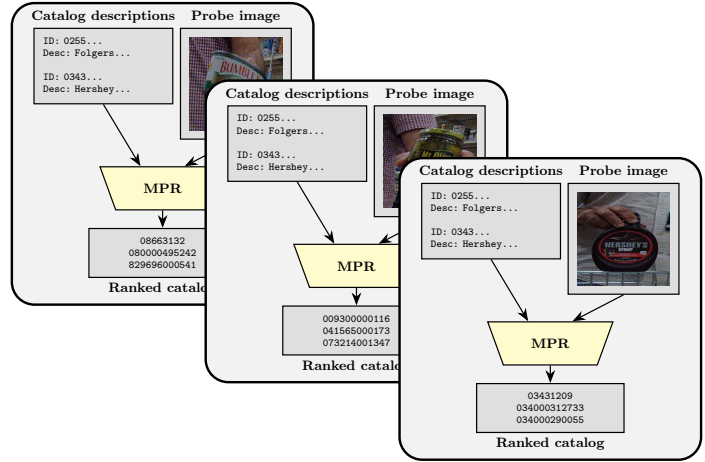
\begin{figure}[!htbp]
    \centering
    \resizebox{0.750\linewidth}{!}{%
    \begin{tikzpicture}[
        card/.style={
            draw,
            ultra thick,
            rectangle,
            rounded corners=15pt,
            fill=white, 
            minimum width=7cm,
            minimum height=7.0cm 
        },
        innerbox/.style={
            draw,
            thick,
            rectangle,
            align=left,
            font=\scriptsize\ttfamily,
            inner sep=6pt,
            fill=white
        },
        mprbox/.style={
            draw,
            thick,
            trapezium,
            trapezium left angle=70,
            trapezium right angle=70,
            shape border rotate=180, 
            minimum width=2cm,
            minimum height=1cm,
            align=center,
            font=\bfseries,
            inner sep=5pt
        },
        arrow/.style={
            ->,
            thick,
            >={Stealth[length=3mm]},
            rounded corners=2pt
        }
    ]

    \newcommand{\drawstage}[5]{
        \begin{scope}[shift={(#2,#3)}]
            
            \node[card, fill=gray!10] (#1_bg) at (0,0) {};

            \node[innerbox, text width=2.8cm, anchor=north west, fill=gray!25] (#1_cat) 
                at ([xshift=-3.0cm, yshift=-0.6cm]#1_bg.north) 
                {ID: 0255...\\Desc: Folgers...\\ \ \\ID: 0343...\\Desc: Hershey...};
            \node[above, font=\small\bfseries] at (#1_cat.north) {Catalog descriptions};

            \node[innerbox, minimum size=1.8cm, anchor=north east, fill=gray!25] (#1_img) 
                at ([xshift=3.0cm, yshift=-0.6cm]#1_bg.north) 
                {%
                 \includegraphics[width=2.15cm, height=2.15cm]{#5}%
                };
            \node[above, font=\small\bfseries] at (#1_img.north) {Probe image};

            \node[mprbox, fill=yellow!25] (#1_mpr) at (0, -0.8) {MPR};

            \node[innerbox, text width=3cm, align=center, fill=gray!25] (#1_rank) 
                at (0, -2.3) 
                {#4};
            \node[below, font=\small\bfseries] at (#1_rank.south) {Ranked catalog};

            \draw[arrow] (#1_cat.south) -- (#1_mpr.north west);
            \draw[arrow] (#1_img.south) -- (#1_mpr.north east);
            \draw[arrow] (#1_mpr.south) -- (#1_rank.north);

        \end{scope}
    }
    
    \drawstage{stage1}{0}{0}{
        08663132\\
        080000495242\\
        829696000541
    }{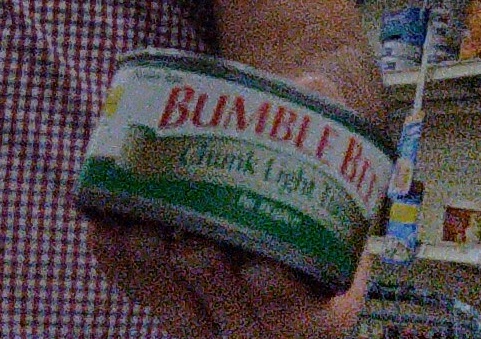}

    \drawstage{stage2}{4.5}{-1.8}{
        009300000116\\
        041565000173\\
        073214001347
    }{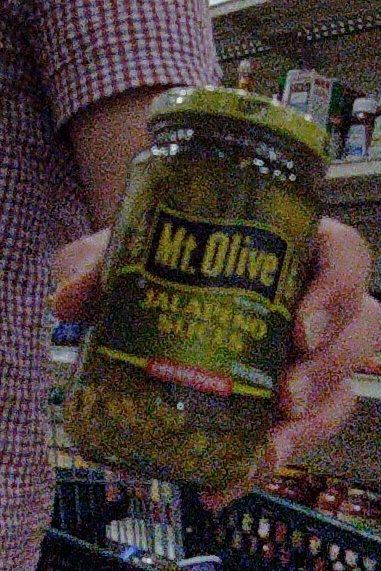}

    \drawstage{stage3}{9}{-3.6}{
        03431209\\
        034000312733\\
        034000290055
    }{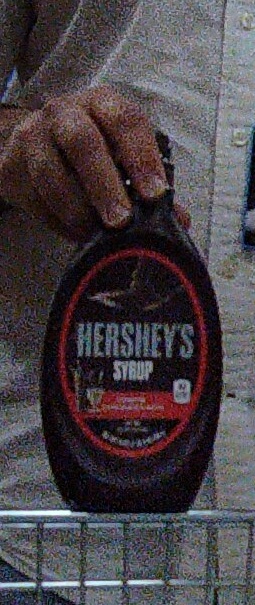}

    \end{tikzpicture}%
    }
    \caption{\textbf{The MPR inference protocol.} MPR frames recognition as a ranking task. For each probe image, cross-modal similarity sorts SKUs by relevance, enabling zero-shot identification.}
    \label{fig:mpr_protocol}
\end{figure}

Visual recognition of packaged goods is the cornerstone of modern retail intelligence, enabling checkout-free stores and automated supply chain management. Historically, this task relied on manual feature extraction before evolving into deep learning models trained for closed-set classification. However, new SKUs and packaging redesigns appear daily, making fixed classifiers operationally unsustainable.

To address this rigidity, the paradigm has shifted toward Multimodal Product Retrieval (MPR) using foundation Vision Language Models (VLMs) such as CLIP~\cite{radford2021learning}. As illustrated in Fig.~\ref{fig:mpr_protocol}, MPR frames recognition as a dynamic ranking task where each probe image is matched against a textual catalog via cross-modal similarity. This approach theoretically enables zero-shot classification but introduces model selection ambiguity. Standard benchmarks like ImageNet prioritize coarse semantic categorization, while retail applications demand fine-grained instance discrimination to separate nearly identical SKUs. Prior works have fine-tuned VLMs for retail domains~\cite{srivastava2023retailklip} or evaluated specific architectures~\cite{czerwinska2025benchmarking}, but no systematic zero-shot benchmark isolates the effects of pre-training data, architecture, and input resolution.

To bridge this gap, we conduct a systematic evaluation of 190 open-source models from the OpenCLIP~\cite{ilharco2021openclip} repository on the 4th GroceryVision Challenge~\cite{groceryvision2025}. We included all dual-encoder checkpoints available as of September 2025, excluding only corrupted weights. Our findings are specific to grocery products, where inter-SKU visual similarity is high; generalization to fashion, electronics, and other retail domains is an open question. Unlike previous studies that conflate multiple variables, we strictly control for architecture, dataset source, and input resolution. Our contributions are:

\begin{enumerate}
    \item \textbf{Data quality trumps scale.} Dataset filtration predicts zero-shot MPR performance better than parameter count. Switching from raw web-scrapes to filtered sources yields up to a 16.6\% accuracy gain, more than doubling the model size would deliver.
    
    \item \textbf{Scaling anomalies favor efficient models.} We identify cases where compact models outperform larger baselines. MobileCLIP-B~\cite{vasu2024mobileclip} (150M parameters) surpasses 351M counterparts trained on unfiltered data. To quantify this, we propose \textit{semantic power density} ($\phi$), an efficiency metric that penalizes poor accuracy-to-parameter ratios.
    
    \item \textbf{A systematic ranking precision gap.} We report that models achieve high Recall@5 (94.5\%) but suffer a 17.5\% drop at Recall@1, indicating that contrastive pre-training produces embeddings suited for coarse retrieval but geometrically unstable for fine-grained ranking.
\end{enumerate}

\section{Related Work}

\paragraph{Product recognition in retail.} Retail product recognition evolved from manual descriptors like SIFT~\cite{lowe2004distinctive} and BRISK~\cite{leutenegger2011brisk} to fine-tuned CNNs~\cite{srivastava2020bag,peng2020rp2k}. Retraining backbones for dynamic catalogs proved prohibitive. GAN-based domain adaptation~\cite{tonioni2019domain} remains bound by closed-set classification rigidity.

\paragraph{Vision-language models and benchmarks.} Foundation models like CLIP~\cite{radford2021learning} established shared embedding spaces enabling open-set retrieval without task-specific training. ELEVATER~\cite{li2022elevater} assesses zero-shot performance across 20 image classification datasets, while DataComp~\cite{gadre2023datacomp} focuses on data-centric evaluation. For fine-grained domains, Products-10K~\cite{bai2020products}, RP2K~\cite{peng2020rp2k}, and GroceryVision~\cite{groceryvision2025} provide SKU-level benchmarks. Our work systematically evaluates zero-shot MPR, isolating pre-training data from architecture.

\paragraph{Efficiency metrics and deployment.} FLOPs and parameter counts remain standard efficiency proxies but fail to capture task-specific utility. Dehghani et al.~\cite{dehghani2021efficiency} show that single cost indicators can mislead and recommend reporting multiple, while scaling-law studies~\cite{hoffmann2022training} characterize compute-optimal model and data scaling. RetailKLIP~\cite{srivastava2023retailklip} fine-tunes a CLIP backbone for retail recognition, and Czerwinska et al.~\cite{czerwinska2025benchmarking} benchmark fine-tuning strategies for e-commerce embeddings. Our semantic power density ($\phi$) introduces a deployment-aware metric penalizing models below functional thresholds.

\section{Preliminaries}
\label{sec:preliminaries}

We define the contrastive loss and ranking metrics used in this benchmark.

\subsection{Contrastive Representation Learning}
We evaluate architectures trained on the CLIP objective, which replaces fixed-label classification with contrastive alignment via InfoNCE loss~\cite{oord2018representation}. The objective maximizes dot-product similarity of $B$ matched image-text pairs $\{(v_i, t_i)\}_{i=1}^B$ while suppressing the $B^2-B$ in-batch negatives. The symmetric loss $\mathcal{L}$ averages image-to-text and text-to-image components.
\begin{equation}
    \mathcal{L}_{I2T} = - \frac{1}{B} \sum_{i=1}^B \log \frac{e^{s(v_i, t_i) / \tau}}{\sum_{j=1}^B e^{s(v_i, t_j) / \tau}}
\end{equation}
\begin{equation}
    \mathcal{L}_{T2I} = - \frac{1}{B} \sum_{i=1}^B \log \frac{e^{s(t_i, v_i) / \tau}}{\sum_{j=1}^B e^{s(t_i, v_j) / \tau}}
\end{equation}
\begin{equation}
    \mathcal{L} = \frac{1}{2}(\mathcal{L}_{I2T} + \mathcal{L}_{T2I})
\end{equation}
Here, $\tau$ is a learnable temperature controlling distribution sharpness. Lower $\tau$ values sharpen the softmax, amplifying gradients from hard negatives. This structures the embedding space such that visual concepts become linearly separable via natural language prompts.

\subsection{Problem Formulation and Metric}
We treat MPR as a ranking problem. Encoders $f_\theta: \mathcal{V} \rightarrow \mathbb{R}^d$ and $g_\phi: \mathcal{T} \rightarrow \mathbb{R}^d$ map images and text onto a shared hypersphere. Relevance is measured by cosine similarity.
\begin{equation}
    s(v, t) = \frac{f_\theta(v) \cdot g_\phi(t)}{\|f_\theta(v)\| \|g_\phi(t)\|}
\end{equation}
For a given query image $v$, we maximize the rank of its ground-truth description $gt_q$. We use the single-gallery-shot protocol where each product has one unique description. The primary metric is Recall@$K$ (CMC at rank $K$), measuring the percentage of queries where the correct match appears in the top $K$ results.
\begin{equation}
    \text{Recall@}K = \frac{1}{Q} \sum_{q=1}^{Q} \mathbb{I}(\text{rank}(gt_q) \le K)
\end{equation}

\section{Benchmark}
We detail the catalog metadata refinement and dual-encoder inference protocol.

\subsection{MPR Dataset}

\begin{figure}[!htbp]
    \centering
    \resizebox{0.750\linewidth}{!}{%
    \begin{tikzpicture}[
        basebox/.style={
            draw,
            rectangle,
            align=center,
            thick,
            inner sep=2mm,
            font=\small
        },
        descbox/.style={
            basebox,
            text width=3.5cm,
            minimum height=1.2cm
        },
        captionbox/.style={
            basebox,
            text width=3.5cm,
            minimum height=2cm
        },
        modelbox/.style={
            basebox,
            text width=4cm,
            minimum height=1.2cm
        },
        arrow/.style={
            ->,
            thick,
            >={Stealth[length=3mm]}
        }
    ]

    
    \node[descbox, fill=blue!25] (orig_desc) {Original description\\($\geq 77$ tokens)};
    
    \node[modelbox, right=0.8cm of orig_desc, fill=orange!25] (llama) {Captioner\\(Llama-3.1-8B-Instruct)};
    
    \node[descbox, right=0.8cm of llama, fill=green!25] (synth_desc) {Synthetic description\\($\leq 77$ tokens)};

    
    \node[captionbox, above=0.8cm of orig_desc, fill=gray!25] (orig_cap) {The image shows a bottle of Hershey's Syrup... \textit{[truncated for brevity]}};
    
    \node[captionbox, above=0.8cm of synth_desc, fill=gray!25] (synth_cap) {The product is Hershey's Genuine Chocolate Syrup, a fat-free, 16 oz dark brown chocolate syrup in a bottle with a black cap.};

    
    \node[basebox, text width=5cm, anchor=south, fill=gray!25] (image_node) at (llama.center |- synth_cap.south) {
        \includegraphics[width=5cm]{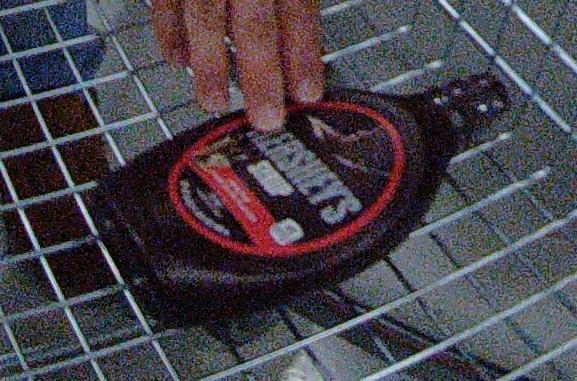} \\ Hershey's Genuine Chocolate Syrup
    };

    \draw[arrow] (orig_desc) -- (llama);
    \draw[arrow] (llama) -- (synth_desc);
    \draw[arrow, <->] (orig_cap) -- (orig_desc);
    \draw[arrow, <->] (synth_cap) -- (synth_desc);

    \end{tikzpicture}%
    }
    \caption{\textbf{Caption refinement.} Llama-3.1-8B-Instruct compresses descriptions exceeding 77 tokens into concise captions while retaining key visual attributes.}
    \label{fig:synth_captioner}
\end{figure}
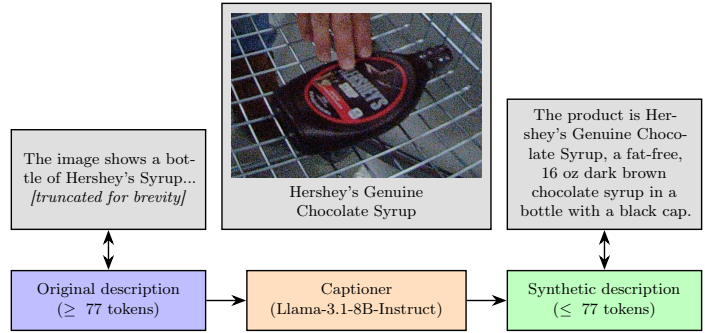

The MPR dataset from the GroceryVision Challenge~\cite{groceryvision2025}, released under CC-BY-NC 4.0 which permits non-commercial use with attribution, contains 74,200 training images across 409 SKUs, subsampled to 12,944 front-facing perspectives. Original catalog metadata frequently exceeds the 77-token context limit of CLIP text encoders, causing truncation of discriminative attributes. We synthesized concise descriptions using Llama-3.1-8B-Instruct~\cite{grattafiori2024llama3herdmodels}. As shown in Fig.~\ref{fig:synth_captioner}, the captioner receives the product description and a tag description emphasizing visual attributes. Outputs begin with \textit{``The product is...''}, prioritize visual features, and remain within 77 tokens. The prompt template is provided below.
\begin{tcolorbox}[colback=blue!5, colframe=blue!90, title=Llama-3.1-8B-Instruct Prompt Template, fonttitle=\bfseries\small, boxrule=0.5pt, arc=2pt]
\small
\textbf{Role:} You are a helpful assistant that generates descriptions for grocery products.

\textbf{Task:} Generate a concise product description in $\leq$77 tokens given the product metadata. Output a JSON object with key \texttt{``label''}.

\textbf{Constraints:}
\begin{itemize}[leftmargin=*, nosep]
    \item Start with ``The product is ...'' in every description.
    \item Prioritize prominent visual attributes from the tag description including color, shape, brand, size, packaging, and form.
\end{itemize}
\end{tcolorbox}

\noindent Generating synthetic descriptions has precedent in retail AI applications~\cite{srivastava2025sgbd}. Crucially, we frame this process as semantic restoration rather than augmentation; it recovers ground-truth visual cues lost in truncated metadata to ensure the retrieval target is factually complete. To validate quality, we randomly sampled 250 descriptions (61\% of the 409-SKU catalog) and reviewed them in two independent passes using different random seeds. Each description was assessed on a pass/fail basis for token compliance and preservation of key visual attributes (color, shape, brand, size) against source metadata. On this sampled subset we observed 100\% token compliance and 100\% attribute retention, with no hallucinations identified.

\subsection{MPR Pipeline}

\begin{figure}[!htbp]
    \centering
    \begin{tikzpicture}[
        scale=0.665, transform shape,
        box/.style={
            draw, thick, rounded corners=4pt, fill=white,
            align=center, font=\small
        },
        component/.style={
            draw, thick, rectangle, rounded corners=6pt,
            minimum width=2.8cm, minimum height=1.2cm,
            font=\bfseries\small,
            align=center,
            fill=white
        },
        embedding/.style={
            rectangle, draw=black!60, thin,
            minimum width=0.4cm, minimum height=1.8cm,
            inner sep=0pt
        },
        mini_embedding/.style={
            rectangle, draw=black!60, thin,
            minimum width=0.4cm, minimum height=0.5cm,
            inner sep=0pt,
            fill=green!30
        },
        scorestrip/.style={
            rectangle, draw=black!30, thin,
            minimum height=0.5cm, 
            font=\tiny\bfseries, text=white,
            anchor=west
        },
        arrow/.style={
            ->, thick, >={Stealth[length=3mm]},
            color=black!80,
            line cap=round
        }
    ]

    \node[box, label=above:{\textbf{Probe image} $v$}, fill=gray!25, align=center] (img_input) at (0, 2.0) {%
    \includegraphics[width=3.2cm, height=3.2cm, keepaspectratio]{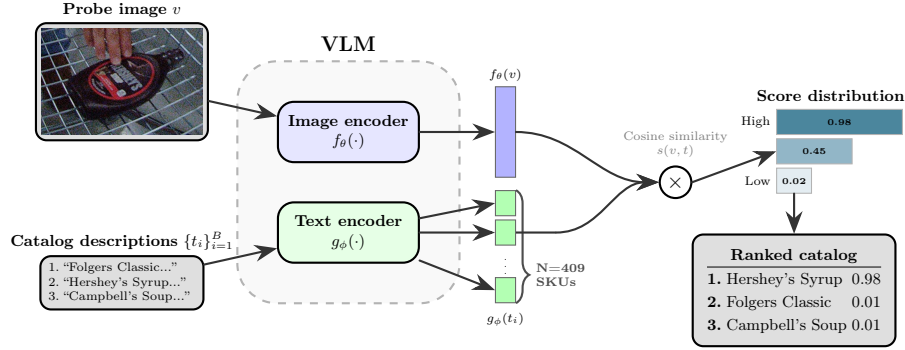}%
    };

    \node[box, label={[align=center]above:\textbf{Catalog descriptions} $\{t_i\}_{i=1}^B$}, fill=gray!25] (txt_input) at (0, -2.0) {
        \scriptsize
        \begin{tabular}{l}
        1. ``Folgers Classic...''\\
        2. ``Hershey's Syrup...''\\
        3. ``Campbell's Soup...''
        \end{tabular}
    };

    \node[component, fill=blue!10] (img_encoder) at (4.5, 1.0) {Image encoder\\$f_\theta(\cdot)$};
    \node[component, fill=green!10] (txt_encoder) at (4.5, -1.0) {Text encoder\\$g_\phi(\cdot)$};

    \draw[arrow] (img_input) -- (img_encoder);
    \draw[arrow] (txt_input) -- (txt_encoder);

    \begin{scope}[on background layer]
        \node[draw=gray!60, dashed, thick, rounded corners=15pt, fill=gray!5, 
              fit=(img_encoder) (txt_encoder),
              inner sep=15pt] (vlm_box) {};
        \node[above=2pt, font=\bfseries\large] at (vlm_box.north) {VLM};
    \end{scope}

    
    \node[embedding, right=1.5cm of img_encoder, fill=blue!30, label=above:\scriptsize $f_\theta(v)$] (img_emb) {};
    \draw[arrow] (img_encoder) -- (img_emb);

    \node[mini_embedding, right=1.5cm of txt_encoder, fill=green!30] (txt_emb2) {}; 
    \node[mini_embedding, above=2pt of txt_emb2, fill=green!30] (txt_emb1) {}; 
    \node[mini_embedding, below=18pt of txt_emb2, fill=green!30] (txt_emb3) {}; 
    
    \node at ($(txt_emb2.south)!0.5!(txt_emb3.north)$) {\tiny $\vdots$};
    \node[below=2pt of txt_emb3, font=\scriptsize] {$g_\phi(t_i)$};

    \draw[arrow] (txt_encoder) -- (txt_emb1.west);
    \draw[arrow] (txt_encoder) -- (txt_emb2.west);
    \draw[arrow] (txt_encoder) -- (txt_emb3.west);

    \draw[decorate, decoration={brace, amplitude=5pt, aspect=0.75}, thick, color=black!70] 
        (txt_emb1.north east) -- (txt_emb3.south east) 
        node[pos=0.75, right=8pt, align=left, font=\scriptsize\bfseries] {N=409\\SKUs};

    \node[circle, draw, thick, fill=white, inner sep=2pt, font=\large] (dotprod) at (11.0, 0) {$\times$};
    \node[above=1.5pt of dotprod, font=\scriptsize, text=gray, align=center] {Cosine similarity\\$s(v, t)$};

    \draw[arrow] (img_emb.east) to[out=0, in=180] (dotprod.west);
    
    \draw[arrow] ($(txt_emb2.east) + (0.8, 0)$) to[out=0, in=180] (dotprod.west);
    \draw[thick, black!80] (txt_emb2.east) -- ($(txt_emb2.east) + (0.8, 0)$);

    \coordinate (score_start) at ($(dotprod) + (2.0, 1.2)$);
    \colorlet{safeTeal}{cyan!50!black}

    \node[above right=0.3cm and -0.5cm of score_start, font=\bfseries\small] {Score distribution};

    \node[scorestrip, fill=safeTeal!90, minimum width=2.5cm, label={[font=\scriptsize]left:High}, text=black] (s1) at (score_start) {0.98};
    \node[scorestrip, fill=safeTeal!50, minimum width=1.5cm, below=2pt of s1.south west, anchor=north west, text=black] (s2) {0.45};
    \node[scorestrip, fill=safeTeal!10, minimum width=0.5cm, below=2pt of s2.south west, anchor=north west, label={[font=\scriptsize]left:Low}, text=black] (s3) {\tiny 0.02};

    \draw[arrow] (dotprod.east) -- (s2.west);

    \node[box, below=0.8cm of s3, fill=gray!25, inner sep=5pt] (topk) {
        \renewcommand{\arraystretch}{1.2}
        \begin{tabular}{@{}l l r@{}}
            \multicolumn{3}{c}{\textbf{Ranked catalog}} \\
            \hline
            \textbf{1.} & Hershey's Syrup & 0.98 \\
            \textbf{2.} & Folgers Classic & 0.01 \\
            \textbf{3.} & Campbell's Soup & 0.01 \\
        \end{tabular}
    };

    \draw[arrow] (s3.south) -- (topk.north);

    \end{tikzpicture}
    \caption{\textbf{Architecture of the MPR pipeline.} The VLM encoders ($f_\theta, g_\phi$) map the probe image $v$ and catalog descriptions $\{t_i\}$ into a shared embedding space. We compute the cosine similarity $s(v, t)$ to rank the 409 SKUs.}
    \label{fig:mpr_pipeline}
\end{figure}

As illustrated in Figure \ref{fig:mpr_pipeline}, the pipeline follows a dual-encoder architecture. The system takes a probe image $v$ and $N=409$ catalog descriptions $\{t_i\}_{i=1}^N$. OpenCLIP instantiates the encoders with diverse checkpoint weights. The image encoder $f_\theta$ and text encoder $g_\phi$ produce $L_2$-normalized embeddings. Relevance is computed via dot product, using sigmoid for SigLIP~\cite{zhai2023sigmoid,tschannen2025siglip2} or softmax for CLIP. All evaluations run on a single A100 (40GB) using \texttt{bfloat16} precision.

\section{Analysis}
\label{sec:analysis}

We evaluate 190 models to isolate the drivers of retrieval performance.

\subsection{Strongest Open Source VLM for MPR}

\textbf{Which pre-trained VLM provides the strongest foundation for MPR?} Our benchmarking identifies SigLIP2~\cite{tschannen2025siglip2} as the strongest foundation. ViT-gopt-16-SigLIP2-384 achieves zero-shot Recall@1 of 0.770, outperforming ViT-L-14~\cite{radford2021learning} by 35\%. All metrics are computed on the full test set with deterministic inference. We verified stability by running three independent forward passes with different batch orderings, observing standard deviations below 0.002 for all Recall@$K$ values. Table \ref{tab:sota_hierarchy} shows modern SigLIP variants dominate the leaderboard. However, state-of-the-art (SOTA) retrieval requires 1.8B parameters. PE-Core~\cite{bolya2025perception} achieves 0.758 with only 671M parameters, offering near-peak performance at one-third the cost.

\begin{table}[!htbp]
\centering
\caption{\textbf{SOTA hierarchy for MPR.} SigLIP2 models establish a new MPR baseline. PE-Core achieves 98\% of peak accuracy at one-third the parameters.}
\label{tab:sota_hierarchy}
\begin{tabular}{lccc}
\toprule
\textbf{Checkpoint} & \textbf{Size (M)} & \textbf{PT data} & \textbf{Recall@1} \\
\midrule
ViT-gopt-16-SigLIP2-384~\cite{tschannen2025siglip2} & 1871 & WebLI-10B~\cite{chen2023pali} & 0.770 \\
PE-Core-L-14-336~\cite{bolya2025perception} & 671 & MetaCLIP-5.4B~\cite{xu2023metaclip} & 0.758 \\
ViT-SO400M-14-SigLIP2-378~\cite{tschannen2025siglip2} & 1136 & WebLI-10B~\cite{chen2023pali} & 0.749 \\
ViT-L-16-SigLIP2-384~\cite{tschannen2025siglip2} & 882 & WebLI-10B~\cite{chen2023pali} & 0.701 \\
\midrule
ViT-L-14~\cite{radford2021learning} & 428 & WIT-400M~\cite{radford2021learning} & 0.568 \\
\bottomrule
\end{tabular}
\end{table}

Our analysis uncovers a non-linear relationship between resolution and accuracy. Table \ref{tab:resolution_wall} shows ViT-L-16-SigLIP2 peaks at 384px and plateaus at 512px. ViTamin-XL~\cite{chen2024vitamin} peaks at 256px with sharp drops at 384px. We hypothesize that beyond optimal resolution, ViTs~\cite{dosovitskiy2020image} over-index on high-frequency noise such as JPEG artifacts and sensor noise. Table \ref{tab:resolution_wall} bears this out: compute grows quadratically while accuracy plateaus or regresses. We conclude 384px is the practical ceiling for MPR.

\begin{table}[!htbp]
\centering
\caption{\textbf{The resolution wall.} Performance plateaus or regresses beyond 384px. Compute cost grows quadratically while accuracy degrades. 384px is the practical efficiency ceiling for MPR.}
\label{tab:resolution_wall}
\begin{tabular}{lcccc}
\toprule
\textbf{Architecture} & \textbf{Low res.} & \textbf{High res.} & \textbf{Delta} & \textbf{Compute cost} \\
\textit{(Comparison)} & \textit{(Recall@1)} & \textit{(Recall@1)} & \textit{(Accuracy)} & \textit{(FLOPs)} \\
\midrule
ViT-L-16-SigLIP2~\cite{tschannen2025siglip2} & 0.701 \scriptsize{(384px)} & 0.700 \scriptsize{(512px)} & \textbf{-0.1\%} & $+1.7\times$ \\
ViTamin-XL~\cite{chen2024vitamin} & 0.687 \scriptsize{(256px)} & 0.670 \scriptsize{(384px)} & \textbf{-2.5\%} & $+2.2\times$ \\
ViT-L-14~\cite{radford2021learning} & 0.568 \scriptsize{(224px)} & 0.561 \scriptsize{(336px)} & \textbf{-1.2\%} & $+2.2\times$ \\
\bottomrule
\end{tabular}
\end{table}

\textbf{What is the best model for a specific memory budget?} Table \ref{tab:size_leaders} shows top models per parameter class. MobileCLIP-B achieves 0.653 Recall@1 despite having half the parameters of ConvNeXt-Large-d-320~\cite{liu2022convnet} at 0.645. The DataCompDR~\cite{vasu2024mobileclip} distillation process explains this. ConvNeXt learns from LAION-2B~\cite{schuhmann2022laion}, a web-scraped dataset with substantial noise from misaligned image-text pairs~\cite{gadre2023datacomp}. MobileCLIP trains on synthetic captions and embeddings distilled from a teacher ensemble pre-trained on filtered data. We did not measure inference latency directly. The $\phi$ metric serves as a theoretical guide but practitioners should validate on target hardware. MobileCLIP-B is optimal for 150M budget on-device applications. PE-Core-L-14 delivers 98.4\% of peak at one-third compute for server-side deployment.

\begin{table}[!htbp]
\centering
\caption{\textbf{Size class leaders.} MobileCLIP-B outperforms larger models through knowledge distillation. Data quality matters more than parameter count for edge deployment. PT: Pre-training.}
\label{tab:size_leaders}
\resizebox{\linewidth}{!}{
\begin{tabular}{llccc}
\toprule
\textbf{Size family} & \textbf{Checkpoint} & \textbf{PT data} & \textbf{Size} & \textbf{Recall@1} \\
\midrule
\textbf{Small} ($<200$M) & MobileCLIP-B~\cite{vasu2024mobileclip} & DataCompDR~\cite{vasu2024mobileclip} & 150M & 0.653 \\
\textbf{Medium} ($200$--$400$M) & ConvNeXt-Large-d-320~\cite{liu2022convnet} & LAION-2B~\cite{schuhmann2022laion} & 351M & 0.645 \\
\textbf{Large} ($400$M--$1$B) & PE-Core-L-14-336~\cite{bolya2025perception} & MetaCLIP~\cite{xu2023metaclip} & 671M & 0.758 \\
\textbf{Very Large} ($>1$B) & ViT-gopt-16-SigLIP2-384~\cite{tschannen2025siglip2} & WebLI~\cite{chen2023pali} & 1.8B & 0.770 \\
\bottomrule
\end{tabular}
}
\end{table}

\subsection{Influence of Pre-training Data on Downstream MPR}

\textbf{What influence do pre-training datasets have on MPR?} Dataset filtration quality determines zero-shot transfer more than model scale. We observe a strict hierarchy where filtered datasets like DataComp~\cite{gadre2023datacomp} consistently outperform raw LAION~\cite{schuhmann2022laion} and WIT-400M~\cite{radford2021learning} across all architectures. Table \ref{tab:universal_data_scaling} shows a monotonic relationship between filtering rigor and Recall@1. The performance gap widens as model capability increases, suggesting higher-capacity models require higher-fidelity signals. Changing data source yields higher return than doubling parameters. Table \ref{tab:noise_tolerance} shows ViT-B-32 suffers 79.4\% failure on noisy CommonPool~\cite{gadre2023datacomp} while ViT-L-14 maintains parity. The large model locks onto the dominant clean signal and ignores outliers. Data purity requirements scale inversely with capacity.

\begin{table}[!htbp]
\centering
\caption{\textbf{Universal scaling of data quality.} We compare Recall@1 of the same architectures pre-trained on different data sources. DataComp-XL (filtered) consistently outperforms LAION-2B (raw scale) and WIT-400M (web scrape), confirming that data quality is a universal efficiency multiplier.}
\label{tab:universal_data_scaling}
\resizebox{\linewidth}{!}{
\begin{tabular}{lcccc}
\toprule
\textbf{Backbone} & \textbf{WIT-400M}~\cite{radford2021learning} & \textbf{LAION-2B}~\cite{schuhmann2022laion} & \textbf{DataComp-XL}~\cite{gadre2023datacomp} & \textbf{Curated gain} \\
\textit{(size)} & \textit{(web scrape)} & \textit{(raw scale)} & \textit{(filtered)} & \textit{($\Delta$ vs WIT-400M)} \\
\midrule
ViT-B-32 (151M)~\cite{radford2021learning} & 0.403 & 0.481 & \textbf{0.521} & +11.8\% \\
ViT-B-16 (150M)~\cite{radford2021learning} & 0.443 & 0.535 & \textbf{0.608} & +16.6\% \\
ViT-L-14 (428M)~\cite{radford2021learning} & 0.568 & 0.619 & \textbf{0.693} & +12.5\% \\
\bottomrule
\end{tabular}
}
\end{table}

\begin{table}[!htbp]
\centering
\caption{\textbf{Capacity works as a filter.} Large models can survive noisy data, while small models collapse. This creates a distillation trap since one cannot simply distill a large, robust model into a small edge model using noisy data. Efficient models require strictly higher-purity data to converge.}
\label{tab:noise_tolerance}
\resizebox{\linewidth}{!}{
\begin{tabular}{lcccc}
\toprule
\textbf{Backbone} & \textbf{WIT-400M}~\cite{radford2021learning} & \textbf{CommonPool}~\cite{gadre2023datacomp} & \textbf{DataComp-XL}~\cite{gadre2023datacomp} & \textbf{Noise penalty} \\
\textit{(size)} & \textit{(baseline)} & \textit{(noisy)} & \textit{(filtered)} & \textit{(CommonPool vs. WIT-400M)} \\
\midrule
ViT-B-32 (151M)~\cite{radford2021learning} & 0.403 & 0.083 & 0.521 & -79.4\% \textit{(collapse)} \\
ViT-B-16 (150M)~\cite{radford2021learning} & 0.443 & 0.267 & 0.608 & -39.7\% \textit{(degraded)} \\
ViT-L-14 (428M)~\cite{radford2021learning} & 0.568 & 0.574 & 0.693 & +1.0\% \textit{(resilient)} \\
\bottomrule
\end{tabular}
}
\end{table}

\textbf{Does architecture protect against data scarcity?} No. YFCC-15M is a 15-million image subset of YFCC100M~\cite{thomee2016yfcc100m} comprising user-uploaded Flickr photos with noisy metadata. Table \ref{tab:scale_collapse} shows ResNet50~\cite{he2016deep} and ResNet101 suffer $>$93\% drops from WIT-400M to YFCC-15M. Collapse is universal across architectures. Consequently, our benchmarking identifies a distinct shift in the data frontier. Table \ref{tab:dataset_leaders} confirms SOTA models abandon raw scraping for filtered data.

\begin{table}[!htbp]
\centering
\caption{\textbf{Universal scale sensitivity.} We compare identical architectures trained on large-scale web data (WIT-400M) vs. small, uncurated subsets (YFCC/CommonPool). The collapse is universal across depth (ResNet101), architecture (ViT), and activation function (quickgelu), confirming that data scale is a primary driver of zero-shot retrieval capability. PT: Pre-training.}
\label{tab:scale_collapse}
\resizebox{\linewidth}{!}{
\begin{tabular}{lccccc}
\toprule
\textbf{Backbone} & \textbf{Large-scale} & \textbf{Small-scale} & \textbf{Recall@1} & \textbf{Recall@1} & \textbf{Collapse} \\
 & \textbf{PT data (400M)} & \textbf{PT data (12--15M)} & \textbf{(Large)} & \textbf{(Small)} & \textbf{ratio} \\
\midrule
ResNet50~\cite{he2016deep} & WIT & YFCC & 0.405 & 0.018 & $-$95.5\% \\
ResNet50-quickgelu~\cite{ilharco2021openclip} & WIT & YFCC & 0.386 & 0.019 & $-$95.1\% \\
ResNet101~\cite{he2016deep} & WIT & YFCC & 0.438 & 0.027 & $-$93.8\% \\
ResNet101-quickgelu~\cite{ilharco2021openclip} & WIT & YFCC & 0.399 & 0.025 & $-$93.7\% \\
ViT-B-32~\cite{radford2021learning} & WIT & CommonPool-S & 0.403 & 0.010 & $-$97.5\% \\
\bottomrule
\end{tabular}
}
\end{table}

\begin{table}[!htbp]
\centering
\caption{\textbf{Dataset leaders.} The top performers all abandon raw scraping. Aggressive filtering or WebLI signals yield the best results.}
\label{tab:dataset_leaders}
\resizebox{\linewidth}{!}{
\begin{tabular}{lllc}
\toprule
\textbf{Data} & \textbf{Methodology} & \textbf{Checkpoint} & \textbf{Recall@1} \\
\midrule
WebLI~\cite{chen2023pali} & Image-text similarity filtering & ViT-gopt-16-SigLIP2-384~\cite{tschannen2025siglip2} & \textbf{0.770} \\
DataComp-1B~\cite{gadre2023datacomp} & Similarity filtering & ViTamin-L2-384~\cite{chen2024vitamin} & 0.697 \\
DataCompDR~\cite{vasu2024mobileclip} & Multi-modal reinforcement & MobileCLIP-B~\cite{vasu2024mobileclip} & 0.653 \\
\midrule
LAION-2B~\cite{schuhmann2022laion} & Raw scaling & ViT-L-14-laion2b\_s32b\_b82k~\cite{ilharco2021openclip} & 0.619 \\
\bottomrule
\end{tabular}
}
\end{table}

\subsection{Quantifying Return on Investment for MPR}

\textbf{Which architecture yields highest ROI?} Standard benchmarks plot accuracy linearly against size, but deployment utility follows a sigmoidal curve. Below 50\% accuracy, a model returns more false charges than correct retrievals (odds ratio $<$ 1) and cannot operate autonomously. Above 80\% accuracy, errors become rare enough to tolerate (odds ratio $\geq$ 4). We therefore anchor our efficiency metric at 50\%, the minimum viability threshold where the odds ratio crosses 1. This utility profile motivates a metric that penalizes sub-threshold models instead of linearly rewarding parameter efficiency. We propose semantic power density ($\phi$) (Eq. \ref{eq:semantic_power}). Drawing from signal processing, we treat the retrieval odds ratio as the signal amplitude, setting $\epsilon = 10^{-6}$. Since effective power scales with the square of the amplitude ($P \propto A^2$), we define the metric as the squared Signal-to-Noise Ratio (SNR) normalized by number of parameters in millions:
\begin{equation}
\label{eq:semantic_power}
  \phi = \frac{\left(\frac{\text{Recall@1}}{1 - \text{Recall@1} + \epsilon}\right)^2}{N_{\text{params}}} \times 100
\end{equation}

\begin{table}[!htbp]
\centering
\caption{\textbf{Comparison of efficiency metrics.} We evaluate three formulations on representative models. The linear metric ($M_1$) incorrectly favors small but inaccurate models. The quadratic-accuracy metric ($M_2$) improves ordering but lacks a principled threshold. Our SNR-based $\phi$ correctly identifies MobileCLIP-B as optimal by severely penalizing sub-50\% models.}
\label{tab:metric_comparison}
\begin{tabular}{lccccc}
\toprule
\textbf{Model} & \textbf{Size (M)} & \textbf{Recall@1} & $\mathbf{M_1}$ & $\mathbf{M_2}$ & $\mathbf{\phi}$ \\
 & & & \scriptsize{(R@1$/N$)} & \scriptsize{(R@1$^2/N$)} & \scriptsize{(SNR$^2/N$)} \\
\midrule
ViTamin-S~\cite{chen2024vitamin} & 62 & 0.432 & \textbf{0.70} & 0.30 & 0.93 \\
ResNet50~\cite{he2016deep} & 102 & 0.405 & 0.40 & 0.16 & 0.45 \\
MobileCLIP-B~\cite{vasu2024mobileclip} & 150 & 0.653 & 0.44 & 0.28 & \textbf{2.37} \\
ViT-L-14~\cite{radford2021learning} & 428 & 0.568 & 0.13 & 0.08 & 0.41 \\
PE-Core-L-14~\cite{bolya2025perception} & 671 & 0.758 & 0.11 & 0.09 & 1.46 \\
\midrule
\multicolumn{3}{l}{\textit{Correct ranking?}} & \textcolor{red}{\ding{55}} & \textcolor{orange}{Partial} & \textcolor{green!60!black}{\ding{51}} \\
\bottomrule
\end{tabular}
\end{table}

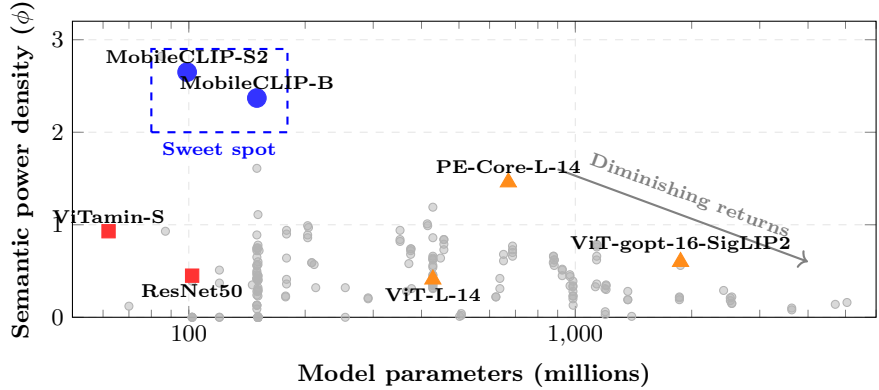
\begin{figure}[!htbp]
    \centering
    \begin{tikzpicture}
    \begin{axis}[
        width=\linewidth,
        height=5.5cm,
        xlabel={Model parameters (millions)},
        ylabel={Semantic power density ($\phi$)},
        xmode=log,
        log ticks with fixed point, 
        ymin=0, ymax=3.2, 
        xmin=50, xmax=6000,
        grid=major,
        grid style={dashed, gray!20},
        ylabel style={font=\small\bfseries},
        xlabel style={font=\small\bfseries},
    ]

    \addplot[
        only marks,
        mark=*,
        mark size=1.5pt,
        color=black!30,
        fill opacity=0.5,
        nodes near coords={},
        point meta=0
    ] coordinates {
        (120, 0.51) (102, 0.45) (102, 0.39) (120, 0.37) (178, 0.23)
        (623, 0.22) (623, 0.22) (178, 0.22) (291, 0.21) (291, 0.20)
        (120, 0.00) (120, 0.00) (102, 0.00) (102, 0.00) (102, 0.00)
        (102, 0.00)

        (151, 1.11) (151, 0.78) (151, 0.60) (151, 0.57) (151, 0.45)
        (151, 0.45) (151, 0.42) (151, 0.40) (151, 0.40) (151, 0.32)
        (151, 0.30) (151, 0.29) (151, 0.28) (151, 0.23) (151, 0.01)
        (151, 0.00) (151, 0.00) (151, 0.00) (151, 0.00) (151, 0.00)
        (151, 0.00) (151, 0.00) (151, 0.00) (151, 0.00) (151, 0.00)
        (151, 0.00) (151, 0.00) (151, 0.00)

        (150, 1.61) (150, 0.89) (150, 0.75) (150, 0.74) (150, 0.72)
        (150, 0.65) (208, 0.59) (208, 0.59) (150, 0.56) (150, 0.56)
        (150, 0.50) (150, 0.46) (150, 0.42) (150, 0.42) (150, 0.41)
        (150, 0.28) (150, 0.24) (150, 0.13) (150, 0.13) (150, 0.13)
        (150, 0.09)

        (428, 1.19) (428, 0.86) (428, 0.66) (428, 0.62) (428, 0.56)
        (428, 0.54) (428, 0.46) (428, 0.45) (428, 0.42) (428, 0.41)
        (428, 0.39) (428, 0.39) (428, 0.38) (428, 0.36) (428, 0.35)
        (428, 0.34) (428, 0.32) (428, 0.31)

        (986, 0.39) (986, 0.34) (986, 0.33) (986, 0.28) (987, 0.23)
        (986, 0.23) (2540, 0.22) (986, 0.20) (1367, 0.19) (1367, 0.19)
        (2540, 0.15) (2540, 0.15) (987, 0.13)

        (203, 0.99) (204, 0.97) (203, 0.89) (203, 0.86) (652, 0.68)
        (878, 0.66) (878, 0.66) (652, 0.61) (877, 0.59) (371, 0.58)
        (1127, 0.33)

        (1136, 0.79) (1137, 0.78) (1136, 0.76) (375, 0.74) (376, 0.74)
        (375, 0.72) (375, 0.68) (1136, 0.66) (882, 0.62) (1135, 0.62)
        (882, 0.62) (1872, 0.60) (882, 0.56) (1871, 0.56) (377, 0.31)

        (62, 0.93) (457, 0.84) (457, 0.84) (457, 0.79) (688, 0.77)
        (151, 0.75) (457, 0.73) (688, 0.73) (688, 0.70) (688, 0.66)
        (211, 0.57) (925, 0.52) (925, 0.51) (925, 0.45) (146, 0.35)

        (179, 0.94) (352, 0.94) (179, 0.90) (352, 0.84) (352, 0.82)
        (152, 0.72) (179, 0.64) (179, 0.41) (179, 0.36) (1201, 0.35)
        (1201, 0.34) (1201, 0.30)

        (414, 1.01) (415, 0.99) (968, 0.46) (969, 0.44) (969, 0.34)
        (2517, 0.21) (2518, 0.18)

        (428, 0.64) (428, 0.63) (150, 0.48) (1136, 0.36) (1367, 0.28)
        (5045, 0.16) (4705, 0.14)

        (1195, 0.06) (507, 0.04) (1195, 0.03) (507, 0.02) (1399, 0.01)
        (502, 0.01)

        (671, 1.46) (87, 0.93) (448, 0.64) (2419, 0.29) (70, 0.12)

        (85, 2.82) (99, 2.65) (150, 2.37) (150, 2.36)

        (638, 0.42) (254, 0.32) (638, 0.31) (254, 0.00)

        (1859, 0.22) (1859, 0.21) (1859, 0.19) (3630, 0.10) (3631, 0.08)

        (213, 0.32) (1193, 0.31) (366, 0.28)
    };

    \pgfplotsset{
        labeled plot/.style={
            scatter,
            only marks,
            visualization depends on={value \thisrow{anchor}\as\myanchor},
            nodes near coords,
            every node near coord/.append style={font=\scriptsize\bfseries, anchor=\myanchor, color=black},
            point meta=explicit symbolic,
            scatter src=explicit symbolic
        }
    }

    \addplot[
        labeled plot,
        mark=square*,
        mark size=2.5pt,
        color=red!80,
    ] table [meta=label, row sep=\\] {
        x      y     label              anchor \\
        62     0.93  {ViTamin-S}        south \\
        102    0.45  {ResNet50}      north \\
    };

    \addplot[
        labeled plot,
        mark=triangle*,
        mark size=3.5pt,
        color=orange!90,
    ] table [meta=label, row sep=\\] {
        x      y     label              anchor \\
        671    1.46  {PE-Core-L-14}     south west \\
        428    0.41  {ViT-L-14}         north \\
        1872   0.60  {ViT-gopt-16-SigLIP2}          south \\
    };

    \addplot[
        labeled plot,
        mark=*,
        mark size=3.5pt,
        color=blue!80,
    ] table [meta=label, row sep=\\] {
        x      y     label              anchor \\
        99     2.65  {MobileCLIP-S2}    south west \\
        150    2.37  {MobileCLIP-B}     south west \\
    };

    \draw[dashed, blue, thick] (axis cs: 80, 2.0) rectangle (axis cs: 180, 2.9);
    \node[blue, font=\bfseries\scriptsize, align=center] at (axis cs: 120, 1.8) {Sweet spot};

    \draw[->, thick, gray] (axis cs: 900, 1.6) -- (axis cs: 4000, 0.6)
        node[midway, above, sloped, font=\bfseries\scriptsize, yshift=1pt] {Diminishing returns};

    \end{axis}
    \end{tikzpicture}
    \caption{\textbf{The edge-centric efficiency landscape.} Semantic power density ($\phi$) vs. model size. \textcolor{gray}{Gray dots} show the full population ($N=190$). \textcolor{blue}{Blue circles} mark high-efficiency MobileCLIP models. \textcolor{red}{Red squares} indicate poor ROI. \textcolor{orange}{Triangles} represent massive models with low density due to parameter bloat. MobileCLIP occupies the sweet spot for edge deployment.}
    \label{fig:phi_vs_params}
\end{figure}

\textbf{Why SNR?} Table~\ref{tab:metric_comparison} shows linear metrics rank ViTamin-S~\cite{chen2024vitamin} (62M, 43.2\%) above MobileCLIP-B~\cite{vasu2024mobileclip} (150M, 65.3\%). Our SNR-based $\phi$ creates a deployment threshold at 50\%. This threshold is principled. At 50\% accuracy, the odds ratio equals 1, meaning success and failure are equiprobable. Below this, the model produces more errors than correct retrievals. The 50\% choice targets autonomous retail systems; applications with cheaper error recovery can re-parameterize $\phi$ without changing its SNR structure. Above 50\%, signal power grows super-linearly, filtering out functionally obsolete architectures (Fig.~\ref{fig:phi_vs_params}). For edge deployment, we propose prioritizing candidates with high Recall@$K$, then validating $\phi$. If a model relies on Recall@5 for coverage, a lightweight reranker bridges the accuracy gap.

\begin{table}[!htbp]
\centering
\caption{\textbf{The efficiency hierarchy.} MobileCLIP-S2 generates 3$\times$ more signal per parameter than ViTamin-S. Reducing size is a false economy if accuracy drops below usable thresholds. \textbf{Bold} is SOTA accuracy. \underline{Underline} is peak density. \underline{\underline{Double underline}} is optimal edge trade-off.}
\label{tab:semantic_density}
\begin{tabular}{lccccc}
\toprule
\textbf{Model checkpoint} & \textbf{Family} & \textbf{Size (M)} & $\mathbf{Recall@5}$ & $\mathbf{Recall@1}$ & $\mathbf{\phi}$ \\
\midrule
\multicolumn{6}{l}{\textit{High density}} \\
MobileCLIP-S2~\cite{vasu2024mobileclip} & MobileCLIP & 99 & 0.835 & \underline{0.619} & \underline{2.65} \\
MobileCLIP-B~\cite{vasu2024mobileclip} & MobileCLIP & 150 & 0.860 & \underline{\underline{0.653}} & \underline{\underline{2.37}} \\
\midrule
\multicolumn{6}{l}{\textit{Low density}} \\
PE-Core-L-14-336~\cite{bolya2025perception} & PE-Core & 671 & 0.937 & 0.758 & 1.46 \\
ViTamin-S~\cite{chen2024vitamin} & ViTamin & 62 & 0.674 & 0.432 & 0.93 \\
ViT-gopt-16-SigLIP2-384~\cite{tschannen2025siglip2} & SigLIP2 & 1871 & \textbf{0.945} & \textbf{0.770} & 0.60 \\
ResNet50~\cite{he2016deep} & ResNet & 102 & 0.637 & 0.405 & 0.45 \\
ViT-L-14~\cite{radford2021learning} & ViT & 428 & 0.814 & 0.568 & 0.41 \\
\bottomrule
\end{tabular}
\end{table}

As shown in Table \ref{tab:semantic_density}, $\phi$ aligns strictly with deployment reality. MobileCLIP-S2 dominates the frontier, generating the highest discriminative signal per unit of hardware. This metric solves the monotonicity problem in linear metrics. ViTamin-S ranks in the low-density tier despite its small size, confirming that parameter savings are outweighed by retrieval failures. The metric also highlights the exponential cost of SOTA. ViT-gopt-16-SigLIP2-384 scores only 0.60, demonstrating that while it provides highest absolute fidelity, its power-to-weight ratio is significantly lower than MobileCLIP variants. Table~\ref{tab:family_breakdown} provides a concise overview of the efficiency landscape aggregated by architectural family.

\begin{table}[!t]
\centering
\caption{\textbf{Efficiency breakdown by family.} Peak $\phi$ across 14 architectural families. MobileCLIP establishes the efficiency ceiling ($\phi=2.82$). Legacy and massive-scale architectures offer the lowest return. PT: Pre-training data.}
\label{tab:family_breakdown}
\resizebox{\linewidth}{!}{
\begin{tabular}{l c c l l c c}
\toprule
\textbf{Family} & \textbf{Count} & \textbf{Max $\phi$} & \textbf{Backbone} & \textbf{PT data} & $\mathbf{Recall@5}$ & $\mathbf{Recall@1}$ \\
\midrule
\multicolumn{7}{l}{\textit{Tier 1: Hyper-efficient ($\phi > 2.0$)}} \\
MobileCLIP~\cite{vasu2024mobileclip} & 4 & \textbf{2.82} & MobileCLIP-S1 & DataCompDR~\cite{vasu2024mobileclip} & 0.818 & 0.608 \\
\midrule
\multicolumn{7}{l}{\textit{Tier 2: High utility ($1.0 \le \phi < 2.0$)}} \\
ViT~\cite{radford2021learning} & 80 & 1.61 & ViT-B-16 & DataComp-XL~\cite{gadre2023datacomp} & 0.816 & 0.608 \\
PE-Core~\cite{bolya2025perception} & 5 & 1.46 & PE-Core-L-14-336 & MetaCLIP~\cite{xu2023metaclip} & 0.937 & 0.758 \\
ViT-CLIPA~\cite{li2023inverse} & 7 & 1.01 & ViT-L-14-CLIPA & DataComp-1B~\cite{gadre2023datacomp} & 0.887 & 0.671 \\
\midrule
\multicolumn{7}{l}{\textit{Tier 3: Moderate density ($0.5 \le \phi < 1.0$)}} \\
ViT-SigLIP~\cite{zhai2023sigmoid} & 11 & 0.99 & ViT-B-16-SigLIP & WebLI~\cite{chen2023pali} & 0.805 & 0.587 \\
ConvNeXt~\cite{liu2022convnet} & 12 & 0.94 & ConvNeXt-Base-w & LAION-2B~\cite{schuhmann2022laion} & 0.773 & 0.565 \\
ViTamin~\cite{chen2024vitamin} & 15 & 0.93 & ViTamin-S & DataComp-1B~\cite{gadre2023datacomp} & 0.674 & 0.432 \\
ViT-SigLIP2~\cite{tschannen2025siglip2} & 15 & 0.79 & ViT-SO400M-14-SigLIP2-378 & WebLI~\cite{chen2023pali} & 0.923 & 0.749 \\
EVA~\cite{sun2023evaclip} & 7 & 0.64 & EVA02-L-14 & Merged-2B~\cite{sun2023evaclip} & 0.859 & 0.624 \\
ResNet~\cite{he2016deep} & 16 & 0.51 & ResNet101 & WIT-400M~\cite{radford2021learning} & 0.684 & 0.438 \\
\midrule
\multicolumn{7}{l}{\textit{Tier 4: Low density or diminishing returns ($\phi < 0.5$)}} \\
CoCa~\cite{yu2022coca} & 4 & 0.42 & CoCa-ViT-L-14 & LAION-2B~\cite{schuhmann2022laion} & 0.822 & 0.621 \\
Roberta-ViT~\cite{liu2019roberta,radford2021learning} & 3 & 0.32 & RoBERTa-ViT-B-32 & LAION-2B~\cite{schuhmann2022laion} & 0.701 & 0.453 \\
ViT-Worldwide~\cite{chuang2025metaclip2} & 5 & 0.22 & ViT-H-14 & Worldwide~\cite{chuang2025metaclip2} & 0.877 & 0.671 \\
NLLB-CLIP~\cite{visheratin2023nllb} & 6 & 0.06 & NLLB-Large-SigLIP & NLLB-v1~\cite{visheratin2023nllb} & 0.714 & 0.453 \\
\bottomrule
\end{tabular}
}
\end{table}

\subsection{Retrieval Performance and the Discriminative Gap}

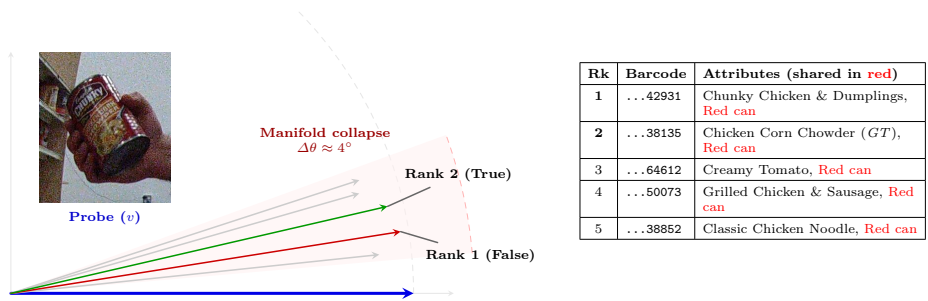
\begin{figure}[!htbp]
    \centering
    \resizebox{\linewidth}{!}{%
    \begin{tikzpicture}[
        lbl/.style={align=left, font=\bfseries\scriptsize, inner sep=1pt}
    ]
        \begin{scope}[local bounding box=plotarea, scale=7.5, cap=round, >=stealth]
            \draw[->, gray!20] (0,0) -- (1.1,0);
            \draw[->, gray!20] (0,0) -- (0,0.6);
            \draw[gray!20, dashed] (1,0) arc (0:45:1);
            
            \fill[red!5, opacity=0.6] (0,0) -- (5:1.15) arc (5:20:1.15) -- cycle;
            \draw[red!30, thin, dashed] (5:1.15) arc (5:20:1.15);
            
            \node[inner sep=0pt, anchor=north west] (img) at (0.07, 0.6) {
                \includegraphics[width=2.8cm, height=2.8cm, keepaspectratio]{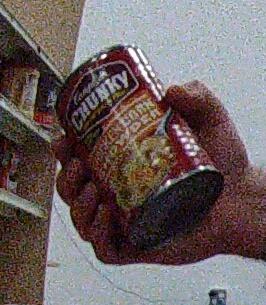}
            };
            \node[below=1pt of img, font=\bfseries\scriptsize, blue!80!black] {Probe ($v$)};
            
            \draw[->, ultra thick, blue!90!black] (0,0) -- (0:1.0);
            \draw[->, thick, gray!40] (0,0) -- (6:0.92);
            \draw[->, thick, gray!40] (0,0) -- (16:0.90);
            \draw[->, thick, gray!40] (0,0) -- (18:0.91);
            \draw[->, thick, red!80!black] (0,0) -- (9:0.98) coordinate (vec_false);
            \draw[->, thick, green!60!black] (0,0) -- (13:0.96) coordinate (vec_true);
            
            \node[coordinate, pin={[pin edge={thick, black!60}, font=\bfseries\scriptsize]330:{Rank 1 (False)}}] at (vec_false) {};
            \node[coordinate, pin={[pin edge={thick, black!60}, font=\bfseries\scriptsize]60:{Rank 2 (True)}}] at (vec_true) {};
            
            \node[align=center, font=\scriptsize, text=red!60!black] at (0.78, 0.38) {\textbf{Manifold collapse}\\$\Delta \theta \approx 4^\circ$};
        \end{scope}

        \node[anchor=west, inner sep=0pt] at ([xshift=0.5cm]plotarea.east) {
            \renewcommand{\arraystretch}{1.4}
            \scriptsize
            \setlength{\tabcolsep}{4pt}
            \begin{tabular}{|c|l|>{\raggedright\arraybackslash}p{4.0cm}|}
            \hline
            \textbf{Rk} & \textbf{Barcode} & \textbf{Attributes (shared in \textcolor{red}{red})} \\ \hline
            \textbf{1} & \texttt{...42931} & Chunky Chicken \& Dumplings, \textcolor{red}{Red can} \\ \hline
            \textbf{2} & \texttt{...38135} & Chicken Corn Chowder (\textit{GT}), \textcolor{red}{Red can} \\ \hline
            3 & \texttt{...64612} & Creamy Tomato, \textcolor{red}{Red can} \\ \hline
            4 & \texttt{...50073} & Grilled Chicken \& Sausage, \textcolor{red}{Red can} \\ \hline
            5 & \texttt{...38852} & Classic Chicken Noodle, \textcolor{red}{Red can} \\ \hline
            \end{tabular}
        };
    \end{tikzpicture}
    }
    \caption{\textbf{Manifold collapse illustrated.} Embedding projection for a representative query. The model must distinguish \textit{Chicken Corn Chowder} from \textit{Chicken \& Dumplings}. Shared visual attributes (\textcolor{red}{red}) cause vectors to collapse into a narrow cone of confusion ($\Delta \theta \approx 4^\circ$). Dot product cannot separate them.}
    \label{fig:degenerate_manifold}
\end{figure}

We evaluate retrieval using Recall@$K$. Table \ref{tab:discriminative_gap} shows ViT-gopt-16-SigLIP2-384 retrieves the correct product within top-5 candidates 94.5\% of the time but fails to rank it first in 17.5\% of cases. Current VLMs operate as category-level retrievers rather than SKU-level rankers. Pre-training on broad alt-text pushes models to ignore subtle details. The result is high Recall@5 but locally degenerate manifold geometry (Fig.~\ref{fig:degenerate_manifold} shows a representative confusion case). Embeddings of distinct SKU variants collapse into a narrow cone around their shared visual attributes. The bottleneck is no longer representational capacity but the ranking mechanism itself.

Our analysis points to a two-stage architecture consisting of a frozen VLM retriever followed by a lightweight reranker. Two-stage and late-interaction ranking are standard in information retrieval~\cite{nogueira2019passage,khattab2020colbert}, and the 17.5\% gap between Recall@5 and Recall@1 quantifies the available headroom. Since the correct SKU is already in the top-5 shortlist 94.5\% of the time, the remaining task is ranking candidates rather than learning representations. Candidate approaches include cross-attention modules on top-$K$ candidates, pairwise ranking transformers, or a compact VLM repurposed for reranking, such as SmolVLM~\cite{marafioti2025smolvlm}.

\begin{table}[!htbp]
\centering
\caption{\textbf{The discriminative gap.} Top models ranked by Recall@5. The gap $\Delta = \text{Recall@5} - \text{Recall@1}$ quantifies precision drop. VLMs excel at narrowing the search space (Recall@5 $\approx 94\%$) but struggle to rank final candidates, with gaps exceeding 17\%.}
\label{tab:discriminative_gap}
\begin{tabular}{lccccc}
\toprule
\textbf{Model checkpoint} & \textbf{Size (B)} & $\mathbf{Recall@5}$ & $\mathbf{Recall@3}$ & $\mathbf{Recall@1}$ & $\mathbf{\Delta}$ \\
\midrule
ViT-gopt-16-SigLIP2-384~\cite{tschannen2025siglip2}  & 1.87 & 0.945 & 0.923 & 0.770 & 0.175 \\
ViT-gopt-16-SigLIP2-256~\cite{tschannen2025siglip2}  & 1.87 & 0.945 & 0.923 & 0.763 & 0.182 \\
PE-Core-L-14-336~\cite{bolya2025perception}      & 0.67 & 0.937 & 0.916 & 0.758 & 0.179 \\
PE-Core-bigG-14-448~\cite{bolya2025perception}   & 2.42 & 0.933 & 0.900 & 0.725 & 0.208 \\
EVA02-E-14~\cite{sun2023evaclip}            & 4.70 & 0.929 & 0.902 & 0.722 & 0.207 \\
\bottomrule
\end{tabular}
\end{table}

\section{Conclusion}
Our benchmark of 190 VLMs is a diagnostic study of multimodal product retrieval. Three findings challenge prevailing assumptions: (i) data quality, not parameter count, determines the zero-shot ceiling (up to 16.6\% gains from filtered sources); (ii) compact models trained on clean data beat noisier larger baselines (MobileCLIP-B at 150M surpasses 351M competitors); (iii) a 17.5\% Recall@5-to-Recall@1 gap shows current VLMs discriminate coarse categories but not SKU-level variants. For deployment, we recommend MobileCLIP-B on edge ($\phi = 2.37$), PE-Core-L-14 on server, and 384px as a practical resolution ceiling.

\paragraph{Limitations.} Our evaluation is confined to the GroceryVision dataset (409 SKUs, 74{,}200 images), a smaller catalog than Products-10K (${\sim}10$K products)~\cite{bai2020products} or RP2K (${\sim}2$K products)~\cite{peng2020rp2k}. Fashion retrieval may call for different resolution ceilings since texture and color demand higher fidelity. Electronics packaging often contains dense text where OCR-augmented models may outperform pure VLMs; we do not claim the 384px ceiling applies universally. We study frozen backbones; fine-tuning and instruction-tuned retrievers are a natural next step.

\paragraph{Future work.} The 17.5\% gap between Recall@5 and Recall@1 directly motivates a two-stage retrieval follow-up: a lightweight cross-attention or pairwise-ranking reranker over top-$K$ candidates, evaluated to see whether patch-level ViT features retain the discriminative signal that pooled contrastive embeddings throw away. Extensions to multilingual catalogs, non-grocery retail domains, and dynamic inventory settings are natural next steps.





%

\subsubsection{Acknowledgements}
We thank the GroceryVision 2025 Challenge organizers for the dataset and the University of the Philippines Diliman for providing the computing resources.

%
%
%
%
{
    \bibliographystyle{splncs04}
    \bibliography{paper}
}
\end{document}